# Mapping the Methodological Space of Classroom Interaction Research: Scale, Duration, and Modality in an Age of AI

Dorottya Demszky[a], Edith Bouton[b], Alison Twiner[c], Sara Hennessy[c], Richard Correnti[d]

[a] Stanford Graduate School of Education, 485 Lasuen Mall, Stanford, CA, 94305

[b] Hebrew University of Jerusalem, Mount Scopus, Jerusalem 9190501, Israel

[c] Faculty of Education, University of Cambridge, 184 Hills Road, Cambridge CB2 8PQ, UK

[d] University of Pittsburgh, Learning Research and Development Center, 3420 Forbes Ave, Pittsburgh, PA 15260

**Acknowledgements:** We thank the participants and organizers of the CEDiR workshop on educational dialogue for the rich discussion that shaped this work. We are especially grateful to Adam Lefstein, Christine Howe, and Julia Snell for sharing their perspectives through the group interview and for reviewing the manuscript. Thanks also to Jim Malamut for helpful conversations and feedback throughout this project.

**ORCIDS:** D.D.: 0000-0002-6759-9367, E.B. 0000-0001-6092-8300 A.T.: 0000-0002-6479-9910, S.H.: 0000-0002-9050-4995, R.C.: 0000-0002-6656-0567

**Conflict of interest:** The authors declare no conflicts of interest.

**Funding:** The research received no external funding.

Correspondence concerning this article should be addressed to Dorottya (Dora) Demszky, 485 Lausen Mall, Stanford, CA 94305. Email: ddemszky@stanford.edu

**Abstract**

Research on classroom interaction has long been divided between large-scale observation and in-depth ethnographic work. We propose a framework mapping this methodological space along three dimensions—scale, duration, and modality—where a study's position shapes what it reveals and obscures. We illustrate it through contrasting studies of dialogic teaching—Howe et al. (2019) and Snell and Lefstein (2018)—and an interview with the lead researchers, organized around three questions: what can be operationalized, what mechanisms become visible, and what translates to practice. We then examine how AI is expanding this space and how the framework can guide research and tool design.

**Introduction**

Teaching quality is among the strongest school-based predictors of student learning (Chetty et al., 2014), but how we study it shapes what we find. Large-scale observation research using standardized instruments has characterized features of instruction and interaction associated with student learning across hundreds of classrooms (Grossman et al., 2013; Hill et al., 2008; Howe et al., 2019; Pianta et al., 2008). Ethnographic and in-depth qualitative work has revealed what those aggregate patterns obscure: who benefits from particular instructional practices, who is excluded, and why interventions that change teacher behavior do not always change what students experience (Lefstein & Snell, 2013; Mehan, 1979; Sedova & Navratilova, 2020). Both traditions have generated substantial evidence, but largely in isolation from one another, leaving the field to accumulate findings without cohering them.

The need for integration is widely recognized (Ercikan & Roth, 2006; Luoto et al., 2025), yet the field lacks a framework for identifying how upstream methodological choices about data collection determine what an approach reveals or obscures. Without such a framework, calls for mixed-methods research remain less informative, urging combination without specifying which combinations address which questions about teaching and learning.

In this article, we propose a framework for mapping the methodological space of classroom research. We argue that studies of classroom interaction vary along three key dimensions — scale, duration, and modality — and that choices along these dimensions jointly determine what a study can and cannot reveal. Traditional approaches cluster at opposite corners of this space, constrained by the historical trade-off between scale and depth. We illustrate the framework through research on dialogic teaching, centering two contrasting studies — Howe et al.'s (2019) large-scale investigation across 72 classrooms and Snell and Lefstein's (2018) year-long linguistic ethnography, repeatedly visiting 3 classrooms. We draw on a group interview in which the lead researchers reflected on what their tradition could and could not see. We then examine how advances in AI are expanding the inhabitable region of the space, and close by considering how the framework can guide future studies and tool design.

## The Methodological Space

Three dimensions capture the choices with the greatest consequences for what classroom research reveals: scale, duration, and modality. We foreground these because they constrain what data exist in the first place — unlike interpretive dimensions such as unit of analysis, or attention to social, historical, and cultural context, or to the substantive content and disciplinary voices in classroom talk, which shape how data are read once collected. These dimensions are theoretically orthogonal (though they covary in practice), and they are the dimensions along which AI is now shifting long-standing constraints. Together, they define a methodological space in which a study's position determines the questions it can ask.

*Scale.* Research on classroom interaction ranges from studies coding tens of thousands of interactions across dozens or hundreds of classrooms to studies tracing a single exchange across multiple lessons. Standardized observation instruments such as CLASS (Pianta et al., 2008), MQI (Hill et al., 2008), PLATO (Grossman et al., 2013) and dialogue-specific schemes such as SEDA (Hennessy et al., 2016) achieve the inter-rater reliability and comparability needed for causal inference and policy guidance, but they do so by decomposing contextualized interaction into pre-specified categories. This often means simplifying the very constructs that matter most. Hennessy et al. (2016), for example, found that moving beyond frequency counts of open and closed questions to code the specific dialogic intention behind communicative acts captured far richer information about classroom dialogue but required interpretive judgments that made reliable coding more difficult. These design choices accumulate. When multiple instruments are applied to the same lessons, they produce markedly different portraits of teaching quality (Klette et al., 2025).

In contrast, discourse-analytic and case-based analyses seek to preserve the sequential, emergent character of instruction as it unfolds (Sherry, 2020). They can trace how a student's contribution reshapes a discussion trajectory, move between grain sizes as needed, and explore what standardized instruments would filter out as analytically central. The limitation is symmetrical. Findings resist comparison across settings and cannot support the statistical inferences needed to inform policy at scale.

*Duration.* Classroom observation research mostly relies on brief windows (one to three observations per teacher, often in a pre-post design) because they are more resource-efficient to capture and enable comparison across many sites (Kane & Staiger, 2012). Such designs can establish that a practice or intervention is associated with an outcome, but they cannot reveal the association's trajectory: whether it persists, how it develops, or why it holds only in some classrooms. They also assume that observed lessons represent typical practice, yet critical features — classroom norms, trust, shared referents that participants draw on across lessons — develop over time and are invisible to a snapshot (Mercer, 2008).

Longitudinal designs open up fundamentally different kinds of inference. With repeated observations, researchers can model the form of change — whether growth in teaching quality is linear, whether it plateaus (Sherry, 2020), and how trajectories vary across teachers, learners and contexts (Briggs & Alzen, 2019). Ethnographic longitudinal work adds a further layer, tracing how meaning is co-constructed cumulatively: how a reference made in one lesson becomes a shared resource later on, how participation identities solidify or shift across an academic year (Mercer, 2008; Twiner et al., 2014). These affordances come at the cost of small samples and limited comparability, but they capture processes that cross-sectional designs structurally cannot see.

*Modality*. The standardized instruments discussed above and dialogic research more broadly rely primarily on verbal behavior, treating speech as the central — often the sole — modality of classroom interaction (Abdu et al., 2021). Transcripts and verbal coding lend themselves to systematic, replicable analysis and are less resource-intensive and more privacy-preserving than video-based alternatives. But analyzing talk alone treats verbal participation as a proxy for engagement, silence as non-participation, and the transcript as a sufficient representation of the interaction.

Multimodal approaches start from the premise that meaning-making is distributed across semiotic modes — gesture, gaze, prosody, and interaction with material or digital objects are constitutive of instruction, not supplementary to it (Goodwin, 2000; Twiner et al., 2021). These approaches can reveal participation that verbal-only methods miss, particularly among students who contribute through embodied rather than spoken means (Yeh et al., 2025) or multimodal dialogue with/within a technology environment (Hennessy, 2025). But multimodal data is more complex and resource-intensive to collect, analyze, and de-identify, and its interpretation can require higher levels of inference compared to verbal interactions where meaning is often more explicit.

**Clusters in opposite corners.** In principle, a study can occupy any position in this three-dimensional space. In practice, resource constraints push research toward two corners. Large-scale observation studies (Howe et al., 2019; Kane & Staiger, 2012; LaRusso et al., 2024) achieve breadth by coding verbal behavior in brief observation windows across many classrooms. Ethnographic and discourse-analytic studies (e.g., Bloome, 2005; Snell & Lefstein, 2018; Twiner et al., 2014) achieve depth by tracing multimodal interaction over extended periods in a few sites. Figure 1 plots these clusters. The few illustrative outliers show that some work does depart from these corners, but typically one dimension at a time. For example, Mercer's Thinking Together studies (2008) combined moderate scale with longitudinal tracking but remained verbal; Abdu et al. (2021) conducted multimodal analysis but only of a single lesson sequence. The middle of the space — richly contextual multimodal analysis sustained across many classrooms over extended periods — has remained largely uninhabitable on practical grounds. Whether emerging tools are shifting that boundary is a question we return to after illustrating the framework through a sustained case.

**Figure 1**

*Three-Dimensional Methodological Space of Classroom Interaction Research*

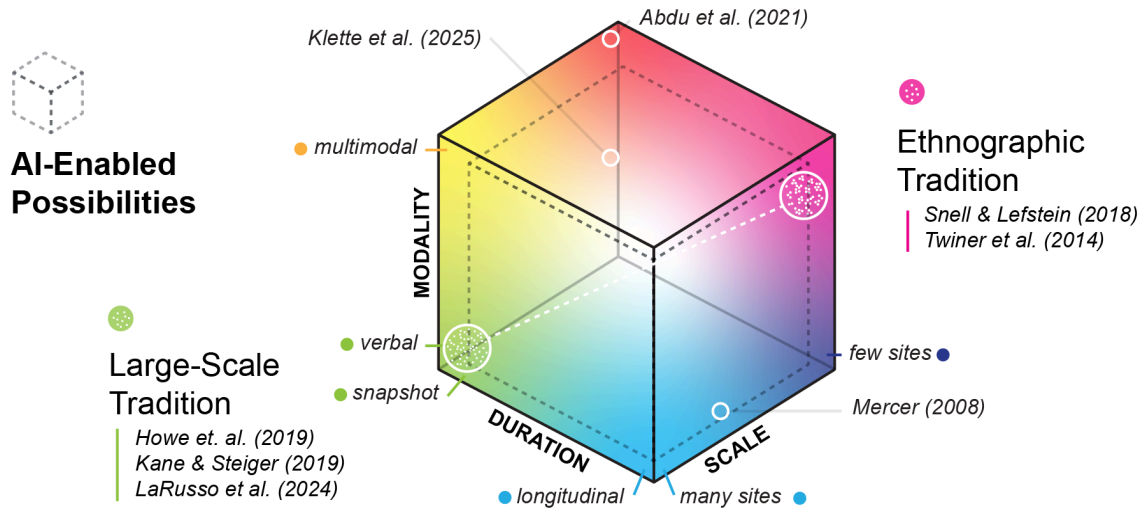

**Illustrating the Framework: Dialogic Teaching as a Case**

    We illustrate the framework through research on dialogic teaching, which positions certain types of classroom talk as central to supporting student thinking and learning (Alexander, 2020). This is a field where the tension between methodological traditions is particularly well-documented (Howe & Abedin, 2013; Kim & Wilkinson, 2019; Lefstein & Snell, 2013), and where the isolation between them has left core epistemological questions unresolved: 1) what can be operationalized as a measurable construct, 2) what mechanisms become visible under different methodological conditions, and 3) how findings translate to practice without losing their meaning. We organize the case around these questions, with each question informing the next, and feeding back into it.

    We center on two studies that occupy opposite corners of the methodological space. Howe et al. (2019) coded every turn across 144 lessons in 72 demographically diverse primary classrooms in England — broad in scale, cross-sectional (two lessons per teacher in different subjects observed in the same week), and primarily verbal. Their analysis examined 33 potential confounds. Snell and Lefstein (2018) conducted linguistic ethnography at one London primary school, including analyzing 71 literacy lessons across three classrooms over a school year — small scale, longitudinal, and multimodal, attending to gaze, gesture, and posture as constitutive of interaction. To push beyond what published texts reveal, we brought the lead researchers, Christine Howe, Julia Snell, and Adam Lefstein, into a group interview structured as a dialogue across traditions, in which each reflected on what their approach could and could not see.

***What can be operationalized?*** Howe et al. operationalized dialogic teaching as 12 codable communicative acts, achieving the reliability needed for statistical inference. But, some constructs became invisible through this decomposition. Coordination of perspectives, where students explicitly connect or reconcile contrasting viewpoints, is one of the most theoretically central features of dialogic pedagogy (Bakhtin, 1981). Yet it appeared roughly 20 times across approximately 80,000 coded turns, too rare to analyze statistically, which Howe described in the interview as "particularly disappointing." Snell suggested studying those rare cases in depth, to "unpick what the conditions are that are facilitating it." Ethnographic research is well-positioned to study how coordination unfolds across turns, which could then reshape what large-scale research measures.

***What mechanisms become visible?*** Both studies considered the role of students' backgrounds in classroom dialogue. Snell and Lefstein traced how teacher perceptions of students' social backgrounds and "ability" shaped access to dialogue: as Julia said, "It wasn't about the background these students came from, it was about how perceptions of that background led to a whole series of events inside the classroom that meant that those students didn't get the chance to participate in dialogue." These events were visible only because the researchers tracked their accumulation over months, attended to gaze and posture as well as talk, and followed individual students rather than sampling across many sites. Howe et al. approached the same question from the opposite corner, treating socioeconomic status as a fixed variable — a predictor and potential confound. Their analysis showed it did not determine dialogue quality; their most dialogically productive classroom was in one of the most deprived areas of the country. Background measured as an administrative variable may not predict dialogue quality, yet still shape it dynamically through teacher perception over time.

***What translates to practice?*** The sharpest disagreement in the interview concerned the move from findings to practical guidance. Howe pointed to a regularity in her team's data: "if the teacher asked for elaboration, overwhelmingly, the next contribution would be an elaboration from the students." If Y almost always follows X, she argued, professional development could simplify the message by focusing on X. Snell pushed back: "You can simplify by saying to teachers, open questions rather than closed questions. But then what is an open question? They ask a question, in its form it seems to be open, but really, that teacher wanted one right answer." Her concern was that large-scale findings, translated into trainable moves, risk producing what Segal and Lefstein (2016) call "exuberant, voiceless participation" — students performing dialogue rituals while the discussion remains fundamentally teacher-directed. Detecting this gap between form and function requires close attention to how episodes unfold, tracked over time, attentive to embodied cues of engagement versus compliance.

These implementation discrepancies close the epistemological cycle. When a practice is adopted in form but not in function, the failure exposes a gap in the original operationalization

— the measurable indicator was a partial proxy for something the design could not reach. Lefstein made this point when Howe instinctively reframed his and Snell's ethnography as a relationship between two variables: "I find it fascinating that when you summarized your understanding of our article, you turned it into a two-variable design, how does one affect the other, which is far from how we understand what we did." The pull toward variable-and-effect thinking, he observed, reshapes how researchers read each other's work and how practitioners interpret findings. On his account, the variables that large-scale research identifies are "indicators of something else [...] a classroom culture, a way of being together." Lefstein speculated that ethnographic understanding of what those indicators actually index could, in turn, inform richer quantitative modeling — looking deeply at cases where Y doesn't follow X, he suggested, could eventually let the field "mathematically model that more complicated picture."

Behavioral regularities identified at scale can be trained, yet performed without producing the phenomenon they are meant to index. When the indicators can be executed without constituting the phenomenon, the definitions need revision. The cycle turns back to the beginning.

## Expanding the Space: AI-Enabled Possibilities

The previous section showed that the epistemological cycle in classroom research turns slowly in part because the methodological space is constrained: the resources required for scale, longitudinal, and multimodal analysis have historically forced researchers into one corner or the other. But advances in AI and natural language processing are beginning to shift these constraints, opening up regions of the methodological space that were previously inaccessible. In this section, we examine what AI makes newly possible along each dimension, what it cannot yet do, and what this means for the epistemological cycle.

***Scale: fine-grained coding across multiple classrooms.*** Automated classifiers make it feasible to operationalize constructs that were previously out of reach at scale, applying fine-grained, turn-level, contextualized coding schemes across thousands of transcripts. The first generation of work, built on smaller, fine-tuned language models, showed that such systems can identify dialogic moves defined by communicative function, which resist keyword- or syntax-based detection such as teachers' focusing questions (Alic et al., 2022) and accountable talk moves (Suresh et al., 2022). These models are typically fine-tuned on thousands to tens of thousands of labeled utterances and then used to predict codes for unlabeled data. Generative large language models (LLMs) such as ChatGPT and Claude extend this capacity in two ways. They can reason over longer stretches of dialogue, making it feasible to code moves whose meaning depends on prior turns. Because they follow codebook instructions with a handful of examples, the challenge shifts from assembling labeled data to articulating the construct (Long et al., 2024; Tran et al., 2024).

Beyond scaling the application of a predefined codebook, LLMs can also serve as collaborators in codebook development itself. They can generate theory-aligned codebooks when prompted with named frameworks that expert coders then refine (Barany et al., 2024), retrieve examples that warrant closer qualitative attention (Katz et al., 2026), and serve as an additional "coder" in calibration sessions, where human–LLM disagreements become occasions for sharpening construct definitions (Ashraf et al., 2026).

Because these tools annotate every turn rather than aggregate to the lesson level, they also support the study of mechanisms by making the interactional conditions surrounding a construct observable — for example, identifying when an authentic question is followed by student reasoning. They can also facilitate translation to practice by providing automated feedback to teachers on their classroom discourse (Demszky et al., 2025; Jacobs et al., 2025; D. Wang, Bian, et al., 2024); by helping curate discourse-based artifacts — short excerpts, annotated exchanges — for use in instructional coaching and PD (Malamut et al., 2025); and by scaffolding teachers' own noticing of dialogic moves, for instance by contrasting their self-identified moves with those flagged by an automated tool (Ashraf et al., 2026).

***Duration: analyzing change over time.*** Longitudinal analysis of classroom observations has historically been constrained not only by the cost of coding but by upstream costs of collecting data — transcription, speaker attribution, and file management across months. Advances in recording hardware and software, automated speech recognition, and speaker diarization now make continuous measurement across dozens of classrooms over a school year feasible, even via teacher self-recording on a phone (Demszky et al., 2025). Yet empirical work remains dominated by snapshot designs (Song et al., 2019; D. Wang, Tao, et al., 2024), leaving substantial room for exploration.

Where scale yields frequency, duration yields form: whether a teacher's use of uptake grows linearly, plateaus, or shifts after a coaching cycle; whether improvements persist, fade, or transfer across units; and how trajectories differ across classrooms. Jacobs et al. (2024), for example, found that teachers who engaged consistently with automated feedback over 2 years increased their accountable talk moves, although change was neither linear nor uniform, corroborating prior work on teacher growth trajectories (Correnti et al., 2021). Denser within-teacher sampling also separates between- from within-teacher variance, making tractable questions of stability, context sensitivity, and the "dose" of a practice needed to matter for outcomes. Longitudinal coding further opens directional questions (does extended student reasoning follow teacher uptake, or precede it?), and supports the study of how students' explanations, arguments, and disciplinary language evolve over time, not just teachers' talk (Correnti et al., 2021). These richer temporal patterns across many classrooms can also help researchers locate the cases that warrant closer interpretive attention.

Duration's most immediate contribution is to the study of mechanisms: it uniquely reveals how conditions at one point shape outcomes at another. But it also reshapes operationalization, since some constructs, like the accumulation of shared classroom norms or the way classroom talk builds on prior lessons and shared texts, are only visible over time. And it matters for translation to practice: interventions require decisions about timing, frequency, and spacing (e.g., coaching cycles calibrated to how long shifts typically take to consolidate), and evaluating them requires measurement that captures persistence and transfer.

*Modality: capturing embodied interaction.* Multimodal analysis of classroom interaction — attending to gesture, gaze, posture, spatial configuration, and prosody alongside speech — has been among the most resource-intensive forms of classroom research. New observation tools (Yeh et al., 2025), computer vision, audio analysis, and multimodal language models are beginning to change this. Although current capabilities are uneven across modalities and tasks, progress has been rapid, with continuous performance improvements and now multimodal language models accepting audio and video directly.

Automated detection of coarse activity configurations — whole-class instruction, small-group work, individual work, transitions — now reaches usable accuracy on full-length classroom video (Foster et al., 2024), and audio pipelines segment and transcribe teacher speech well when a dedicated microphone is available (D'Mello et al., 2015). Other signals are harder: student speech is difficult to segment without per-student microphones (J. Wang et al., 2024), fine-grained interpretive labels (e.g., the function of a gesture or the stance conveyed by prosody) still depend on human judgment (Foster et al., 2024), and multimodal language models struggle to accurately locate specific moments in long classroom videos (Lee et al., 2026). Thus, while AI can now help scale the identification of classroom configuration, participation, and movement, it cannot yet replace human analysis of gesture, stance, and meaning.

Modality's most immediate contribution is to operationalization: embodied constructs like joint attention, uptake, and participation can now be measured at scale. It also opens mechanisms, since the links between teachers' positioning, gaze, and gesture and students' engagement are hard to study from transcripts alone. And it matters for translation to practice: video has long been used to support teacher reflection and coaching because watching teaching is closer to experiencing teaching than reading transcripts (Tao et al., 2026), but curating useful clips is labor-intensive — a bottleneck multimodal AI could assist with.

*Blind spots and ethics.* Most language models were not pretrained on classroom data and generalize unevenly to subjective, context-dependent constructs and systematically under-detect rare or culturally specific phenomena, creating a structural bias toward what is frequent and familiar — what Snell called the "blinkering" risk that coded categories pull attention toward whatever the scheme already recognizes. Models can also be swayed by features that should be irrelevant: LLMs evaluating instructional quality shift their judgments in response to spurious social context such as teacher demographics or experience level (Nam & Demszky, 2026). And they can identify that a teacher asked an authentic question or took up a student idea, but not whether this was the pivotal moment of the lesson or whether it was handled well; recognizing pedagogical excellence still requires a human eye. Unlike humans, models produce labels without reliable reasoning, making it hard to tell whether a system is tracking the construct or a correlated surface feature, as Lefstein noted.

Ethical considerations are integral (cf. Alwahaby & Cukurova, 2024). Each step up the modality ladder (transcripts, audio, video) raises the data's identifiability, and continuous recording with automated analysis can slip from research tool to monitoring apparatus, especially where teachers and students are already over-surveilled. Data collected for research or coaching should not drift into evaluation or disciplinary decisions; consent structures and institutional safeguards need to hold that line. Performance disparities across dialects, accents, and student populations can themselves encode bias, so deployment without auditing these disparities risks amplifying inequity rather than reducing it.

These limits shape how AI should – and should not – be used in classroom research. AI expands the region researchers can inhabit, but the tensions the framework identifies persist. The ethnographic corner — building relationships with participants, understanding the social and cultural meaning of classroom life, recognizing what is significant about an anomaly — is not a bottleneck AI removes but a form of knowledge production with its own epistemological commitments, and those commitments are what allow it to generate insights large-scale research cannot. Humans also remain central to construct articulation, interpretation, and judgments of quality, and to improving the tools themselves. When human and model disagree, humans decide whether the codebook or prompt needs refinement, or whether the model needs fine-tuning. The corners will stay, but now the field can move more easily between them.

**Looking Ahead**

The framework is meant to be prospective. Before a study begins, it lets researchers ask what a given position along scale, duration, and modality will allow them to see, and what it will necessarily miss. Single-tradition studies remain essential, but the field would benefit from research programs that move between corners rather than waiting for traditions to converge. AI-enabled broad tracking that surfaces cases warranting close interpretive attention is one such design; others remain to be articulated, and the framework can help specify what each would and would not reveal.

The same logic extends from research designs to the tools they produce. Data collection pipelines, observation rubrics, and teacher-facing feedback systems all encode implicit positions in the methodological space, and the choice of position shapes what they reward or render invisible. When those positions go unstated, the methodological choices inside these tools tend to be read as neutral technical defaults, and what a given tool happens to measure quietly becomes the field's operative definition of good teaching. The form-versus-function gap surfaced by Snell and Lefstein is one consequence: a tool measuring dialogue's surface features without the means to detect when they have been performed without substance will entrench that gap rather than close it. The most useful thing the framework can do, in our view, is make these positions explicit — pointing toward tools that serve the full epistemological cycle, not only applying a construct but surfacing the cases where it breaks down, so that what counts as a productive dialogic move can be revised rather than fixed.